\documentclass{article}

\usepackage{arxiv}

\usepackage[utf8]{inputenc} % allow utf-8 input
\usepackage[T1]{fontenc}    % use 8-bit T1 fonts
\usepackage{hyperref}       % hyperlinks
\usepackage{url}            % simple URL typesetting
\usepackage{booktabs}       % professional-quality tables
\usepackage{amsfonts}       % blackboard math symbols
\usepackage{nicefrac}       % compact symbols for 1/2, etc.
\usepackage{microtype}      % microtypography
\usepackage{graphicx}
\usepackage{natbib}
\usepackage{doi}
\usepackage{xcolor}         % colors

\usepackage{amsmath}
\usepackage{amssymb}
\usepackage{mathtools}
\usepackage{amsthm}
\usepackage{subcaption}
\usepackage{multirow}

\usepackage{algorithm}
\usepackage{algorithmic}

\usepackage{tikz}
\usetikzlibrary{arrows.meta}
\usetikzlibrary{decorations.pathreplacing,decorations.pathmorphing}
\usetikzlibrary{shadows}

\newtheorem{theorem}{Theorem}
\newtheorem{proposition}{Proposition}

\title{Structured Transforms \\for Low-Overhead Quantization of Language Models}

\author{%
  Daria Cherniuk\thanks{Corresponding author.} \\
  \texttt{kamikazizen@gmail.com} \\
  \And
  Alexander Rudikov \\
  Institute of Numerical Mathematics \\
  \And
  Boris Kashin \\
  Steklov Mathematical Institute
  \And
  Ivan Oseledets \\
  Institute of Numerical Mathematics \\
}

\renewcommand{\shorttitle}{Structured Transforms for Low-Overhead Quantization}

\hypersetup{
pdftitle={Structured Transforms for Low-Overhead Quantization of Language Models},
pdfsubject={cs.LG, cs.CL},
pdfauthor={First Author, Second Author, Third Author},
pdfkeywords={Post-training quantization, Large language models, Kashin decomposition, Discrete cosine transform, Greedy algorithms},
}

\begin{document}
\maketitle

\begin{abstract}
We revisit Kashin-decomposition-based weight quantization for large language models and propose an improved algorithm with stronger convergence properties and structured, efficient orthogonal transforms.
The method retains the core factorization of each weight into two components -- one with bounded infinity norm and the other with bounded infinity norm after an orthogonal transformation -- but replaces the dense random orthogonal matrix with a sign-randomized Discrete Cosine Transform (DCT), reducing the per-iteration cost from $\mathcal{O}(N^2)$ to $\mathcal{O}(N \log N)$.
The proposed greedy algorithm with alternating updates guarantees the four-peak distribution required for stable 2-bit clustering of each factor and admits closed-form initialization of cluster centers, removing the multi-restart k-means bottleneck of prior work.
Composed with OPTQ-style sequential error compensation and QuIP-style incoherence preprocessing, the resulting JAX pipeline is competitive with OPTQ, QuIP, QuIP-RG and a fine-tuning- and vector-quantization-free variant of QuIP\# at 4-bit per channel on OPT, Llama-2 and Pythia, with favorable wall-clock scaling.
The bounded-$\ell_\infty$ factorization is also notably robust: on stress configurations where QuIP variants diverge to four-digit perplexity (Pythia-6.9B) or abort with NaNs in LDL back-substitution (Mistral-7B), Kashin-DCT remains numerically stable and stays close to FP16 baseline.
At inference time, each weight decomposes into two 2-bit factor codes per channel that are structurally suited to native-2-bit hardware.
\end{abstract}

% keywords can be removed

\section{Introduction}\label{sec:introduction}

Modern Large Language Models (LLMs) achieve their capabilities at the cost of weight tensors that dominate memory bandwidth and footprint at inference time.
Post-training quantization (PTQ) compresses these weights to low bit-widths without re-training, making it the practical choice when the training pipeline and large-scale compute are unavailable.
Lower inference precision also translates into lower energy draw per query and a correspondingly smaller carbon footprint.
In deployment, the dominant scheme remains plain uniform scalar quantization with per-channel or per-group scales~\citep{llmint8}, sometimes with a non-uniform grid such as NF4~\citep{qlora}.
%  -- calibration-free, trivial to implement, and easy to fuse into matmul kernels.
The research literature improves on this baseline first via second-order error compensation~\citep{frantar2023optq, OBC} and incoherence preprocessing~\citep{chee2024quip2bitquantizationlarge, quip_better}, and most recently via \emph{vector quantization} (VQ): AQLM~\citep{egiazarian2024extreme}, GPTVQ~\citep{vanbaalen2024gptvq}, and QTIP~\citep{tseng2024qtip} currently lead sub-4-bit benchmarks.
VQ comes with three practical costs, however: each layer must store learned codebook entries alongside the integer indices, the strongest results require gradient-based fine-tuning, and codebook lookups do not fuse into standard matmul kernels as cleanly as scalar dequantization.

A complementary line of work~\citep{merkulov2024quantization} approaches low-bit quantization through \emph{Kashin decomposition}: each vectorized weight matrix $w$ is factorized as $w = u + P^T\hat{v}$, where $P$ is a random orthogonal matrix, $u$ and $\hat{v}=Pv$ separately have small infinity norms.
Empirically, the components produced by Kashin's greedy algorithm form distributions with four sharp symmetric peaks~\citep{merkulov2024quantization}, which is a near-ideal target for 2-bit clustering.
% A 2-bit code per factor then yields an effective 4-bit representation per scalar, but with a dictionary adapted to each weight column rather than a fixed uniform grid.
Despite this attractive structure, the prior application of Kashin decomposition to LLMs~\citep{merkulov2024quantization} suffered from three weaknesses.
First, to amortize cost the authors reformulated the algorithm matrix-wise, which dropped the per-vector convergence guarantees of the original theorem~\citep{kashin1977diameters}; on a non-trivial fraction of layers the iteration failed to converge.
Second, even on layers where the iteration did converge, the four-peak structure that 2-bit clustering relies on was not consistently produced: the joint $u$-$\hat v$ distribution could collapse rather than separate into the four symmetric modes (Figure~\ref{fig:failed_case_opt125m}(a));
Third, the four cluster centers were recovered with multi-restart $k$-means, an unstructured search that dominated quantization wall-clock time.
% and that occasionally produced empty clusters propagating NaNs back through the rotation.

\paragraph{Contributions.}
We revisit Kashin-decomposition-based quantization end-to-end and address all three weaknesses, while keeping the analysis at the vector level where rigorous convergence proofs are available.
\begin{itemize}
    \item \textbf{A partitioned greedy algorithm with guaranteed peaks and a structured DCT transform (Section~\ref{sec:partitioned_greedy_algorithm}).} Building on the recent acceleration of Kashin's theorem in~\citep{kashin2025accelerated}, we propose a Greedy Algorithm with Alternating Updates (Algorithm~\ref{alg:dct_scan_update}) that fixes the order of $u$- and $\hat v$-updates in blocks of four. The schedule guarantees four-peak distributions in both factors, and we prove the conversion guarantees. We further replace the dense random orthogonal $Q$ by a sign-randomized Discrete Cosine Transform with $\mathcal{O}(N \log N)$ cost and zero stored matrix.
    \item \textbf{Closed-form cluster centers (Section~\ref{sec:faster_quantization}).} The greedy updates change the factors by some value $\pm c_k(r_k)$ at each step, so after two updates the four peak locations are exactly $\pm c_1 \pm c_2$, known analytically from the $r_k$ residual norm. We initialize $k$-means with these centers and run only a handful of refinement iterations, eliminating multi-restart search and reducing quantization time.
    \item \textbf{Adaptive integration and a JAX pipeline (Section~\ref{sec:experiments}).} We compose Kashin decomposition with OPTQ-style sequential error compensation and QuIP-style incoherence preprocessing (Hadamard or Kronecker), and implement the full pipeline in compile-friendly JAX with multi-GPU \texttt{pmap} support. On OPT, Pythia, and Llama-family models at 4-bit per channel, the method matches or surpasses OPTQ, QuIP, QuIP-RG, and (fine-tuning-free, vector-quantization-free) QuIP\# on WikiText-2 and C4 perplexity and on HellaSwag, PiQA, and Winogrande accuracy.
\end{itemize}

\section{Related Work}\label{sec:related_work}

Quantization techniques for deep neural networks can be broadly categorized into post-training quantization (PTQ) and quantization-aware training (QAT). QAT typically delivers the highest accuracy because it embeds low-precision constraints directly into the optimization loop; however, it also demands extensive GPU time and memory, especially when working with the very large models that stand to benefit most from reduced-precision inference. 
PTQ, in contrast, converts a pre-trained model to lower bit-widths without additional gradient-based fine-tuning, making it attractive for practitioners who lack the computational resources or training datasets.

The fastest and easiest-to-implement PTQ schemes rely on simple heuristics, such as uniform round-to-nearest or stochastic rounding applied per layer or per row~\cite{llmint8}, or assume a particular activation/weight distribution, as in NormalFloat 4-bit (NF4) quantization~\cite{qlora}. 
Activation- and outlier-aware methods reduce the dynamic range that scalar quantization must cover, either by smoothing the activation/weight magnitudes \citep{xiao2023smoothquant}, by scaling salient channels \citep{lin2024awq}, by jointly learning weight clipping ranges and equivalent transformations \citep{shao2024omniquant}, or by routing outliers to a dense non-uniform codebook \citep{kim2024squeezellm}.

Recent state-of-the-art methods use second-order information to minimize the quantization-induced error in the layer outputs. OPTQ~\cite{frantar2023optq} sequentially quantizes chunks of layer weights while compensating for the accumulated error through corrections to the yet-to-be-quantized parameters of the layer. 
QuIP~\citep{chee2024quip2bitquantizationlarge} shows that OPTQ is a special case of LDLQ adaptive quantization, essentially the method introduced in QuIP without incoherence processing, in which the correction to the yet-to-be-quantized rows of the weight matrix is a linear combination of the quantization errors of the already-quantized rows. Authors further propose incoherence pre- and post-processing to bring weight matrices into the conditions of their theorem, which establishes that LDLQ quantization is a lower bound on nearest and stochastic rounding.

QuIP\#~\citep{quip_better} replaces multiplication by random orthogonal matrices with randomized Hadamard transforms for incoherence processing, which yields better incoherence properties and faster runtime. However, QuIP\# uses vector quantization and resorts to fine-tuning. 
The same randomized-rotation idea is the basis for a parallel line of activation-quantization methods: QuaRot~\citep{ashkboos2024quarot} fuses Hadamard rotations into the residual stream so that both weights and activations are quantized in the rotated basis, SpinQuant~\citep{liu2025spinquant} learns the rotation matrices on the calibration distribution, and DuQuant~\citep{lin2024duquant} composes block-rotations with permutations to suppress outliers further. 
% Our use of a sign-randomized DCT is in the same family of structured-orthogonal-transform methods, but applied to factorize the weight matrix into two bounded-$\ell_\infty$ components rather than to rotate weights and activations into a single low-outlier basis.

Beyond QuIP\#, a substantial line of work pushes weight compression further by switching from scalar to \emph{vector} codebooks. 
AQLM~\citep{egiazarian2024extreme} learns an additive composition of small codebooks per group of weights; GPTVQ~\citep{vanbaalen2024gptvq} extends OPTQ-style error compensation to multi-dimensional codebooks; and QTIP~\citep{tseng2024qtip} replaces QuIP\#'s lattice codebook with trellis-coded quantization on top of the same incoherence processing. These methods achieve very strong sub-4-bit compression ratios, but share two practical costs: each layer must store learned codebook entries alongside the integer indices, and reaching the reported quality typically requires gradient-based fine-tuning or extensive calibration sweeps. The Kashin decomposition we revisit in this work, by contrast, retains \emph{scalar} 2-bit clusters and incurs only a handful of centroids per column (Section~\ref{sec:faster_quantization}), with no fine-tuning step.

% However, these and other compensation-based teqniques~\citep{zhao2025benchmarkingposttrainingquantizationllms} process quantization sequentially, which increases compression time linearly with the number of compressed layers. 

\begin{figure*}[t!]
\normalsize
\centering
    \begin{center}
        \moveright  10pt \hbox{\input{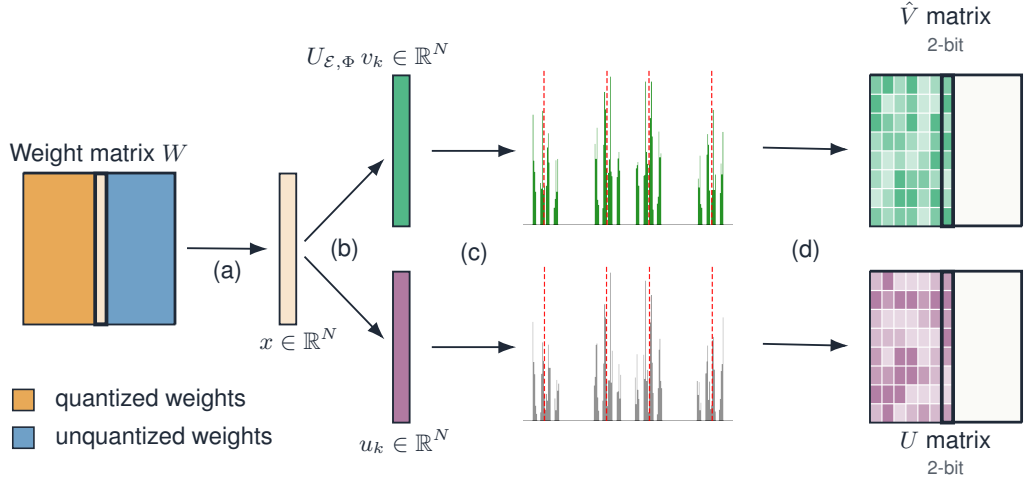}}
    \end{center}
\caption{The proposed quantization pipeline consists of several stages: (a) move to the next column to quantize; (b) decompose with the proposed Greedy Algorithm with Alternating Updates (Algorithm~\ref{alg:dct_scan_update}) into $u$ and $\hat{v}$; (c) quantize each factor to 2 bits via 4-peak clustering; (d) stack the quantized vectors into the output matrices $U$ and $\hat{V}$; then add error compensation to the unquantized weights and return to (a).}
  \label{fig:scheme_offline}
\vspace*{4pt}
\end{figure*}

\section{Problem Setting}\label{sec:problem_setting}

The work of \cite{kashin1977diameters} has introduced the following theorem.
\begin{theorem}
\label{thm:kashin_original}
    For each $N=2,3,\dots$ and for each orthogonal transformation $P \in \mathbb{O}^N$, with the exception of a set $V \subset \mathbb{O}^N, \quad \mu_H(V) \leqslant 2^{-N}$,
    the following inequality holds:
    \begin{equation}
        \label{eq:main_theorem}
        \max\big\{\|x\|_{1}, \,\|Px\|_{1}\big\} \geqslant c_1 \cdot\sqrt{N} \|x\|_2 \qquad \forall x \in \mathbb{R}^N,
    \end{equation}
    where $c_1 > 0$ is an absolute constant.
\end{theorem}
Using \eqref{eq:main_theorem}, for each $P \notin V$, a greedy algorithm~\citep{temlyakov2011greedy} was constructed such that, for every $x \in B_2^N = \{x \in \mathbb{R}^N: \Vert x \Vert_2 \leqslant 1\}$, after $k$ steps, it produces vectors $u_k$ and $v_k$ that satisfy
\begin{equation}
    \label{eq:inifinity_uppder_bounds}
    \quad \Vert u \Vert_{\infty} \leqslant \dfrac{c_2}{\sqrt{N}}, \,\,\Vert Pv \Vert_{\infty} \leqslant \dfrac{c_2}{\sqrt{N}},
\end{equation}
\begin{equation*}
    \big\Vert x - u_k - v_k \big\Vert_{2} \leqslant \gamma^k,
\end{equation*}
where $\gamma < 1$ and $c_2$ are absolute constants. See Appendix~\ref{app:kashin} for full algorithm.
The algorithm requires storing $N^2$ entries of the matrix $Q$ and entails a computational complexity of $\mathcal{O}(N^2)$ operations per iteration.

The upper bounds on the infinity norm in \eqref{eq:inifinity_uppder_bounds} naturally lead to using this factorization for quantization of neural network weights, since the most common uniform quantization suffers severely from outliers~\citep{Nagel2021AWP}.
An even more interesting phenomenon is that the distributions of both $u_k$ and $Pv_k$ often form four distinctive symmetric peaks~\citep{merkulov2024quantization}, which enables straightforward clustering-based quantization.
The former work used this factorization to quantize the weights of an LLM.
Their approach reformulated Kashin's theorem for the matrix case but failed to provide a rigorous convergence analysis for that reformulation.
Consequently, the approach encountered convergence failures and ill-defined peaks on a fraction of layers.
Moreover, the clustering process relied on the $k$-means algorithm with multiple restarts, which significantly slowed quantization.

In this work, we retain the vector formulation, for which rigorous proofs are available, while alleviating the large size of the matrix $Q$ noted in~\cite{merkulov2024quantization} by using a sign-randomized Discrete Cosine Transform (DCT) as the orthogonal transformation, leveraging the recent acceleration of \cite{kashin2025accelerated}; we then partition the greedy schedule so that the dominant updates land at analytically known centroid locations, which removes the multi-restart $k$-means step. The proposed Greedy Algorithm with Alternating Updates and its convergence analysis are presented in Section~\ref{sec:partitioned_greedy_algorithm}; the closed-form $k$-means initialization derived from this schedule is described in Section~\ref{sec:faster_quantization} below.
We further combine Kashin quantization with adaptive quantization methods that quantize model weights chunk by chunk, compensating for the introduced error by adaptively updating the yet-to-be-quantized weights, in the spirit of OPTQ and QuIP; experimental details are provided in Section~\ref{sec:experiments}.

\begin{figure}[t!]
  \centering
        \includegraphics[width=0.99\textwidth]{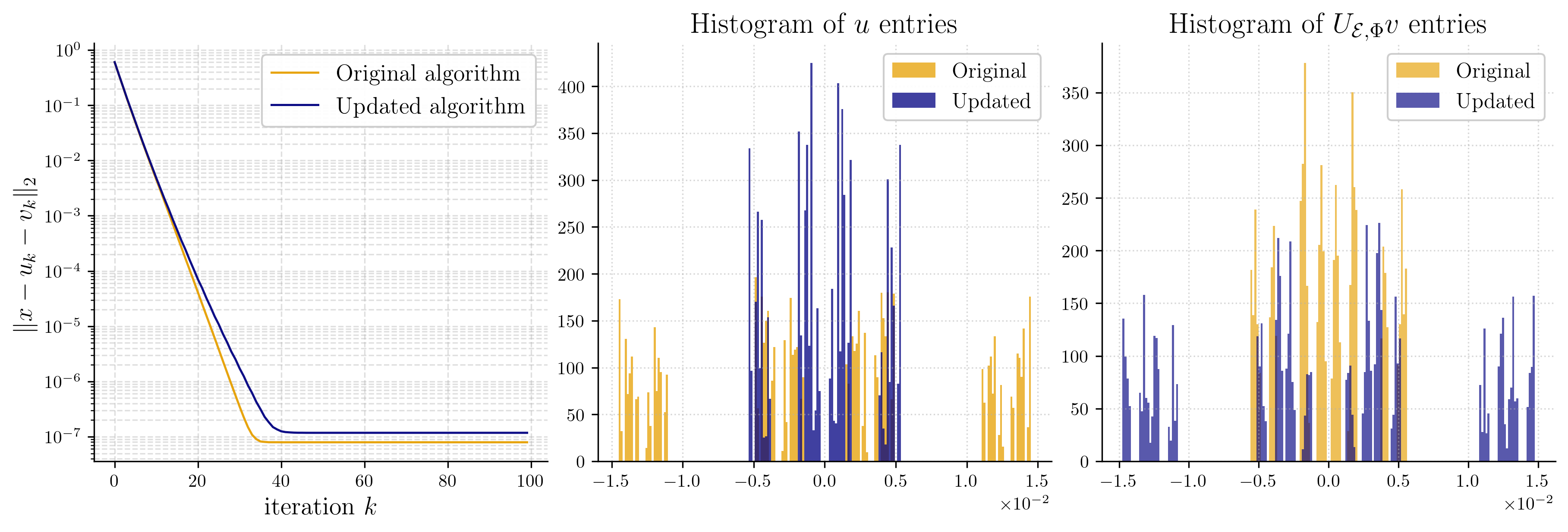}
  \caption{Convergence and factor-distribution comparison on a single random vector $x \in \mathbb{R}^N$ with $N = 10^4$ drawn from $\mathcal{N}(0, I_N)$. Left: residual norm $\bigl\|x - u_{k} - v_{k}\bigr\|_2$ versus the number of greedy updates for the original greedy algorithm of \cite{kashin1977diameters} (Algorithm~\ref{alg:kashin_vec} in Appendix~\ref{app:kashin}, violet) and our partitioned variant of Algorithm~\ref{alg:dct_scan_update} (yellow). Center, right: empirical densities of $u_k$ and $U_{\mathcal{E},\Phi}\,v_k$ at convergence; the partitioned schedule produces four sharp symmetric peaks, while the original schedule leaves the second factor unimodal.}
 \label{fig:convergence}
\end{figure}

\subsection{Closed-form $k$-means initialization}\label{sec:faster_quantization}

Previous work~\citep{merkulov2024quantization} applied $k$-means clustering to the four-peak distributions of $u$ and $Pv$ with random centroid initialization and multiple restarts ($n_{\mathrm{init}} = 50$ in their reported configuration), which dominated quantization wall-clock time.
We exploit the structure of the updates in Algorithm~\ref{alg:kashin_vec}: at iteration $k$, the update of the vector $u$ is $\tfrac{1}{N}\|r_k\|_1 \operatorname{sign}(r_k)$, which corresponds to adding $\pm c_k(r_k)$ where $r_k$ is a residual at iteration $k$.
% with $c_k = \|r_k\|_1/N$ at every coordinate.
After two such updates, the marginal distribution of $u$ contains four peaks at exactly $\pm c_1 \pm c_2$, known analytically from the residual norms.
Because the residual norm decreases rapidly across iterations, these early updates already provide accurate approximations of the peak locations in the final distribution; in practice the first four iterations are sufficient.
The fixed schedule of Algorithm~\ref{alg:dct_scan_update} (Section~\ref{sec:partitioned_greedy_algorithm}) ensures that exactly two updates per factor land at the analytic centroids.
% , simplifying a branch-free, JIT-compilable JAX implementation.
After decomposition converges, the analytic $\pm c_1 \pm c_2$ values are used to initialize the cluster centers and a small number of $k$-means refinement iterations are run; no multi-restart outer loop is required.
The quantization procedure for $\hat{v}$ follows the same strategy.
The closed-form initialization removes the multi-restart loop entirely, reducing the clustering wall-clock time by roughly $10\times$ relative to the multi-restart 2-D $k$-means baseline of \citep{merkulov2024quantization} on medium-sized models.
% Empty-cluster events, which can otherwise propagate NaN values through the inverse rotation, are handled by retaining the previous centroid for any cluster that empties on a given iteration.

\section{The Greedy Algorithm with Alternating Updates}\label{sec:partitioned_greedy_algorithm}

\subsection{Background}
Let $\Phi = \{\phi_j\}_{j=1}^{N}$ be an orthonormal basis in $\mathbb{R}^N$ such that $\Vert \phi_j \Vert_{\infty} \leqslant  K \big/ \sqrt{N},\, 1 \leqslant j \leqslant N.$
For the orthonormal DCT-II used throughout this paper\footnote{Throughout this paper we use the orthonormal DCT-II (the \texttt{norm='ortho'} convention), under which $P_{\mathrm{IDCT}} = P_{\mathrm{DCT}}^\top$}, $\phi_j(n) = \sqrt{2/N}\cos\bigl(\pi(2n+1)j/(2N)\bigr)$ for $j = 0,\dots,N-1$, so $\|\phi_j\|_\infty \leqslant \sqrt{2/N}$ and the constant $K$ in our analysis equals $\sqrt 2$, i.e.\ a small absolute constant independent of $N$.
% \begin{equation*}
%     \label{eq:basis}
%     \big\Vert \phi_j \big\Vert_{\infty} \leqslant \dfrac{K}{\sqrt{N}}, \quad 1 \leqslant j \leqslant N.
% \end{equation*}
As in \cite{kashin2025accelerated}, we consider the orthogonal operator $U_{\mathcal{E}, \Phi} = \mathcal{F}_{\Phi}^{\,-1}\,T_{\mathcal{E}}\, \mathcal{F}_{\Phi},$
% \begin{equation*}
%     \label{eq:operator}
%     U_{\mathcal{E}, \Phi} = \mathcal{F}_{\Phi}^{\,-1}\,T_{\mathcal{E}}\, \mathcal{F}_{\Phi},
% \end{equation*}
where the orthogonal operator $\mathcal{F}_{\Phi}$ acts in $\mathbb{R}^N$ according to the rule $\mathcal{F}_{\Phi}\big(\sum_{j=1}^{N}a_j\phi_j\big) = \{a_j\}_{j=1}^{N} = a \in \mathbb{R}^N,$
% \begin{equation*}
%     \mathcal{F}_{\Phi}\Big(\sum_{j=1}^{N}a_j\phi_j\Big) = \big\{a_j\big\}_{j=1}^{N} = a \in \mathbb{R}^N,
% \end{equation*}
let $\mathcal{F}_{\Phi}^{\,-1}$ be the inverse of $\mathcal{F}_{\Phi}$, and let the operator $T_{\mathcal{E}}$ be defined for a given random set of signs $\mathcal{E} = \{\varepsilon_j\}_{j=1}^{N}, ~\varepsilon_j = \pm 1, ~1 \leqslant j \leqslant N$ by the relation $T_{\mathcal{E}}(a) = \{\varepsilon_j a_j\}_{j=1}^{N}.$
% \begin{equation*}
%     T_{\mathcal{E}}(a) = \big\{\varepsilon_j a_j\big\}_{j=1}^{N}.
% \end{equation*}

In \cite{kashin2025accelerated}, the authors leveraged the fact that for most sign sets $\mathcal{E}$, the following relation holds:
\begin{equation}
    \label{eq:kashin_rudikov}
    \max \Big(\big\Vert x \big\Vert_{1},\,\big\Vert {U_{\mathcal{E}, \Phi}}x \big\Vert_{1}\Big) \geqslant \dfrac{R(N)}{K\sqrt{2}} \qquad \forall x \in \mathbb{R}^{N},\,\,\big\Vert x \big\Vert_{2} = 1,
\end{equation}
where $R(N) = \dfrac{\sqrt{N}}{c_3 \big(\log N\big)^{1/2}\big(\log \log N\big)^{3}}$, and $c_3$ is an absolute constant.
Throughout the rest of the paper we instantiate $\mathcal{F}_\Phi$ as the orthonormal DCT-II and write $P = U_{\mathcal{E},\Phi}$; with $\mathcal{F}_\Phi^{-1} = \mathcal{F}_\Phi^\top$ the operator $P$ admits the FFT-fast realization $Pz = \mathrm{IDCT}\bigl(\varepsilon \odot \mathrm{DCT}(z)\bigr)$ used in Algorithm~\ref{alg:dct_scan_update}. 
% We choose the DCT-II over a randomized Hadamard transform because it is defined for arbitrary $N$, whereas Hadamard requires power-of-two dimensions and is therefore awkward for the heterogeneous hidden sizes of modern LLMs.

The authors of~\cite{kashin2025accelerated}, using the result from \cite{guedon2008}, propose a greedy algorithm in $N$-dimensional Euclidean space with the dictionary $S = Q_N ~\bigcup ~U_{\mathcal{E}, \Phi} Q_N,$
% \begin{equation*}
%     S = Q_N ~\bigcup ~U_{\mathcal{E}, \Phi} Q_N,
% \end{equation*}
where $Q_N \subset \mathbb{R}^N$ is the set of vertices of the cube $B^N_{\infty}$ (i.e., the set of vectors of the form $(\delta_1, \ldots, \delta_N)$ with $\delta_i = \pm 1, ~1 \leqslant i \leqslant N$). It is shown that if \eqref{eq:kashin_rudikov} holds for a sign set $\mathcal{E}$, then for an arbitrary vector $x \in \mathbb{R}^N$ with $\Vert x \Vert_2 \leqslant 1$, this greedy algorithm constructs vectors $u_k$ and $v_k$ from $\mathbb{R}^N$ in $k$ steps such that:
\begin{equation}
    \label{eq:kashin_theorem}
    \begin{split}
        &\big\Vert x - u_k - v_k\big\Vert_{2} \leqslant \big(1 - \alpha(N)\big)^{k/2}, \qquad  \alpha(N) = \dfrac{R^2(N)}{2K^2N}\\
        &\max \Big(\big \Vert u_k \big \Vert_{\infty},\,\big\Vert U_{\mathcal{E}, \Phi}v_k\big\Vert_{\infty}\Big) \leqslant \dfrac{4c_3^2K^2\log N \big(\log\log N\big)^6}{\sqrt{N}}. 
    \end{split}
\end{equation}
In this paper, based on numerical experiments, we propose a modification of the algorithm from \cite{kashin2025accelerated} that offers several advantages over the original method. Notably, while the modification provides practical benefits, the resulting theoretical bound on the convergence rate for the new algorithm is slightly weaker than the estimate provided in \cite{kashin2025accelerated}.

\begin{figure}[t!]
  \centering
  \includegraphics[width=0.99\textwidth]{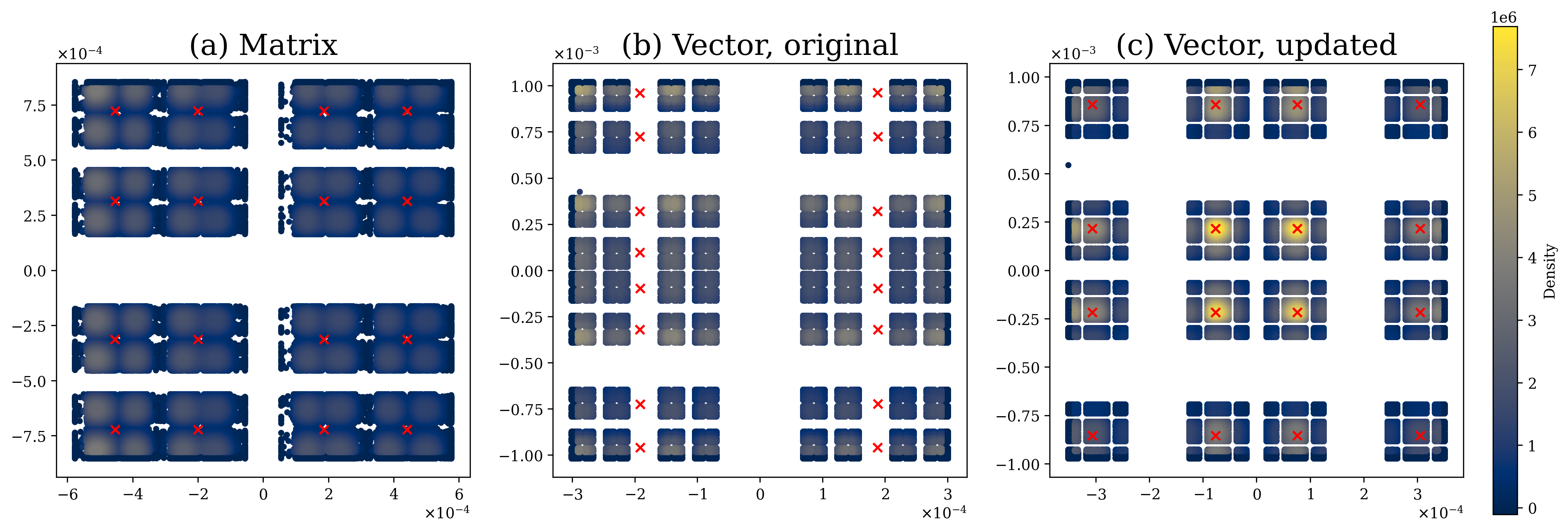}
  \caption{Joint value distribution of $u$ and $\hat{v}$ for the fc2 layer weights of the third decoder block of OPT-125m, decomposed by three algorithms. Red crosses indicate the cluster centers found by $k$-means after factorization.
  \emph{(a) Matrix reformulation of \cite{merkulov2024quantization}}: the matrix-wise greedy iteration drops the per-vector convergence guarantee and the $u$-$\hat v$ joint distribution collapses, leaving cluster centers ill-defined.
  \emph{(b) Original vector greedy algorithm \citep{kashin1977diameters} (Algorithm~\ref{alg:kashin_vec})}: per-vector convergence is restored but the $\hat v$-axis remains spread-out, so 2-bit clustering of $\hat v$ is degenerate.
  \emph{(c) Proposed Algorithm~\ref{alg:dct_scan_update}}: the partitioned schedule and DCT-based $P$ produce four-peaked marginals along both axes, yielding well-separated cluster centers.}
  \label{fig:failed_case_opt125m}
\end{figure}

\subsection{Proposed Algorithm}
The proposed modification retains the dictionary
$S = Q_N \cup U_{\mathcal{E},\Phi}Q_N$ of~\cite{kashin2025accelerated},
but replaces the single-step global greedy selection with a \emph{partitioned} scheme that alternates between the two halves in blocks of four greedy steps: within each block, the first two atomic greedy steps draw from $Q_N$ (and accumulate into $u$), while the last two draw from $U_{\mathcal{E},\Phi}Q_N$ (and accumulate into $\hat v = U_{\mathcal{E},\Phi}\,v$); see Algorithm~\ref{alg:dct_scan_update}.
The slowdown $\beta(N) = \alpha^2(N)/36$ established in Proposition~\ref{main_proposition} is the explicit price of fixing this schedule rather than letting each step adaptively choose the better half as in~\cite{kashin2025accelerated}; in exchange, the dominant updates of $u$ and $\hat v$ land at the analytic centroid locations $\pm c_1 \pm c_2$ used by the closed-form $k$-means initialization of Section~\ref{sec:faster_quantization}. By symmetry of the convergence argument the reverse schedule $(U_{\mathcal E,\Phi}Q_N, U_{\mathcal E,\Phi}Q_N, Q_N, Q_N)$ 
yields the same residual norm at every block of 4 iterations. 
% and our implementation uses the reverse order for kernel-fusion convenience.

\begin{algorithm}[tb]
   \caption{Greedy Algorithm with Alternating Updates (DCT-based realization).}
   \label{alg:dct_scan_update}
\begin{algorithmic}[1]
   \STATE \textbf{Input:} Vector $x \in \mathbb{R}^N$, sign mask $\varepsilon \in \{-1,+1\}^N$, number of blocks $k$
   \STATE \textbf{Output:} Vectors $u_k, \hat{v}_k \in \mathbb{R}^N$ such that
   \[
      x \;\approx\; u_k + P\hat{v}_k \;=\; u_k + v_k,
   \]
   where the orthogonal operator $P:\mathbb{R}^N\to\mathbb{R}^N$ is defined by
   \[
      P z \;:=\; \mathrm{IDCT}\,\big(\varepsilon \odot \mathrm{DCT}(z)\big).
   \]
   \STATE Initialize $u_0 \gets \mathbf{0}_N,\ \hat{v}_0 \gets \mathbf{0}_N,\ r_0 \gets x$
   \FOR{$t = 0,\,1, \dots, k-1$}
      \STATE $\rho \gets r_t,\quad u \gets u_t,\quad \hat v \gets \hat{v}_t$
      \FOR{$j = 1,\,2$}
         \STATE \textit{// two greedy steps from $Q_N$}
         \STATE $\Delta u \gets \mathrm{sign}(\rho)\cdot\dfrac{\|\rho\|_1}{N}$
         \STATE $u \gets u + \Delta u,\qquad \rho \gets \rho - \Delta u$
      \ENDFOR
      \FOR{$j = 3,\,4$}
         \STATE \textit{// two greedy steps from $U_{\mathcal{E},\Phi}Q_N $}
         \STATE $\Delta\hat v \gets \left(\mathrm{sign}(P \rho)\cdot\dfrac{\|P \rho\|_1}{N}\right)$
         \STATE $\hat v \gets \hat v + \Delta\hat v,\qquad \rho \gets \rho - P\,\Delta\hat v$
      \ENDFOR
      \STATE $u_{t+1} \gets u,\quad \hat{v}_{t+1} \gets \hat v,\quad r_{t+1} \gets \rho$
   \ENDFOR
   \STATE \textbf{Return:} $u_k,\ \hat{v}_k,\ r_k$ (and $v_k = P\hat{v}_k$ if needed, by symmetry $P^2 = I$)
\end{algorithmic}
\end{algorithm}

\subsection{Convergence Analysis}\label{sec:convergence_analysis}

We now estimate the convergence rate of Algorithm~\ref{alg:dct_scan_update}. Proposition~\ref{main_proposition} below shows that one block of four atomic greedy steps contracts the residual norm by a factor $(1 - \beta(N))^{1/2}$ with $\beta(N) = \alpha^2(N)/36$; iterating yields the geometric residual decay of Theorem~\ref{proposed_theorem}. 
Reaching a target residual norm therefore takes polylogarithmic factor of iteration steps of the unconstrained algorithm of~\cite{kashin2025accelerated}.
% Proofs are deferred to Appendix~\ref{app:proof_proposition} (with the residual-after-three-steps inequality \eqref{eq:res_3} as the key intermediate bound).

\begin{proposition}
    \label{main_proposition}
    Let $\alpha(N) = R^2(N)\big/2K^2N$ and $\beta(N) = \alpha^2(N)\big/{36}$.
    For any $x \in \mathbb{R}^N$ with $\|x\|_2 \leqslant A$,
    one block of the partitioned greedy algorithm produces a residual $r_1$
    satisfying
    \begin{equation}
        \label{eq:proposition}
        \| r_1 \|_2 \leqslant A\big(1-\beta(N)\big)^{1/2}.
    \end{equation}
\end{proposition}
\begin{theorem}
    \label{proposed_theorem}
    Let $\alpha(N) = R^2(N)\big/2K^2N$ and $\beta(N) = \alpha^2(N)\big/36$.
    For any $x \in \mathbb{R}^N$ with $\|x\|_2 \leqslant 1$,
    the partitioned greedy algorithm after $k$ blocks of iterations constructs
    $u_k, v_k \in \mathbb{R}^N$ such that
    \begin{equation}
        \label{eq:theorem}
        \begin{split}
            \bigl\|x - u_k - v_k\bigr\|_2
                &\leqslant \big(1-\beta(N)\big)^{k/2}, \\[6pt]
            \max\Bigl(
                \bigl\|u_k\bigr\|_{\infty},\,
                \bigl\|U_{\mathcal{E},\Phi}\,v_k\bigr\|_{\infty}
            \Bigr)
                &\leqslant \dfrac{4}{\beta(N)\sqrt{N}}
                    \;=\; \dfrac{c_4\,K^4\,(\log N)^{2}\,(\log\log N)^{12}}{\sqrt{N}},
        \end{split}
    \end{equation}
    where $c_4$ is an absolute constant.
\end{theorem}

% \begin{remark}
% The $\ell_\infty$ bound above follows from a coefficient-summation argument that does not exploit the fine $\ell_1$-structure of the residual. A Kashin-type sharpening following the argument of~\cite{kashin2025accelerated} would replace the right-hand side with $c_\varepsilon K^4 (\log N)^{2+\varepsilon}/\sqrt N$ for any $\varepsilon > 0$. We do not reproduce that sharpening here; for the $N \approx 10^3$--$10^4$ regime relevant to LLM hidden dimensions the looser bound stated above is already informative.
% \end{remark}

\paragraph{Proof sketch.}
The inequality~\eqref{eq:kashin_rudikov} guarantees that, for any residual $\rho$, at least one of the two dictionary halves $Q_N$ or $U_{\mathcal{E},\Phi}Q_N$ is the ``good'' half, in the sense that a single greedy step from it contracts the squared $\ell_2$ norm by a factor $(1-\alpha(N))$. The unconstrained algorithm of \cite{kashin2025accelerated} picks the good half adaptively at every step. 
The partitioned algorithm cannot: it is forced to spend steps $1,2$ on $Q_N$ regardless. 
Proposition~\ref{main_proposition} reduces to two cases. \textbf{(i)} If $Q_N$ \emph{is} the good half for the block's input residual $\rho_0$, step $1$ alone already gives $\|\rho_1\|_2 \leqslant A\sqrt{1-\alpha}$, which is stronger than the claim; steps $2$--$4$ are non-expansive. \textbf{(ii)} If $U_{\mathcal{E},\Phi}Q_N$ is the good half, steps $1,2$ on $Q_N$ are ``wasted'' but, by orthogonality of consecutive greedy updates, their combined contribution has norm $\leqslant A\sqrt{2\beta(N)}$. Step $3$ from $U_{\mathcal{E},\Phi}Q_N$ is then at least as good as the best one-shot approximation $\lambda^* w^*$ of $\rho_0$ from that half, which by the one-step estimate satisfies $\|\rho_0 - \lambda^* w^*\|_2 \leqslant A\sqrt{1-\alpha}$. The triangle inequality combines these into
$\|\rho_3\|_2 \leqslant A\sqrt{1-\alpha} + A\sqrt{2\beta} \leqslant A\sqrt{1-\beta}$,
where the final step is exactly the algebraic identity that fixes the constant $\beta(N) = \alpha^2(N)/36$. Theorem~\ref{proposed_theorem} then iterates Proposition~\ref{main_proposition} over $k$ blocks for the $\ell_2$ bound, and bounds $\|u_k\|_\infty$ and $\|U_{\mathcal{E},\Phi}v_k\|_\infty$ by summing the per-step coefficients $|\lambda_j^{(\ell)}| \leqslant \|r_{j-1}\|_2/\sqrt{N}$ across blocks; the geometric residual decay makes that sum a convergent series of order $1/(\beta(N)\sqrt{N})$.
Full proofs of Proposition~\ref{main_proposition} and Theorem~\ref{proposed_theorem} can be found in Appendix~\ref{app:proof_proposition}.
\section{Experiments}\label{sec:experiments}

We conduct experiments on the OPT \citep{zhang2022opt}, Llama-2 \citep{touvron2023llama2}, Mistral \citep{jiang2023mistral7b}, and Pythia \citep{biderman2023pythia} families of models.
We use WikiText-2 \citep{merity2017pointer} and C4 \citep{raffel2020t5} for measuring perplexity, and HellaSwag \citep{zellers2019hellaswag}, PiQA \citep{bisk2020piqa}, and Winogrande \citep{sakaguchi2021winogrande} for zero-shot accuracy.
All evaluations are run through the \texttt{lm-evaluation-harness} framework\footnote{\url{https://github.com/EleutherAI/lm-evaluation-harness}} for reproducible and reliable results, with maximum sequence length set to 2048.
For all adaptive methods, the per-layer Hessian $H = X^\top X$ is estimated on $1000$ sequences of length $2048$ drawn from the WikiText-2 (\texttt{wikitext-2-raw-v1}) training split, and the same calibration set is used across all baselines. 
% The Hessian update block size~\citep{frantar2023optq} is set to $128$ and Hessian damping to $\lambda = 10^{-2}\,\mathrm{tr}(H)/N$ (\texttt{percdamp}$=0.01$), applied identically across methods.
Unless stated otherwise, each row in Table~\ref{tab:results} reports mean $\pm$ standard deviation over three random seeds; the seed determines the calibration data shuffle for all methods, the sign set $\varepsilon$ for our Kashin-DCT pipeline, and the random incoherence rotation for any method that uses one.
All quantization and evaluation runs are performed on a single NVIDIA H100 GPU; 
per-method end-to-end quantization wall-clock times are reported in Appendix~\ref{app:runtime}, and an inference-time analysis for the Kashin representation is given in Appendix~\ref{app:inference}.

First, we demonstrate the advantages of our Greedy Algorithm with Alternating Updates over the original greedy algorithm.
Figure~\ref{fig:convergence} shows that, on a random vector drawn from $\mathcal{N}(0, 1)$, Algorithm~\ref{alg:dct_scan_update} attains nearly the same convergence rate as the original vector greedy algorithm while producing markedly sharper peak definition.

Second, we show that our method does not fail on layers where the previous Kashin-decomposition-based approach encountered problems.
Figure~\ref{fig:failed_case_opt125m} compares the joint $u$ and $\hat{v}$ distributions for the matrix reformulation of~\cite{merkulov2024quantization}, the original greedy algorithm, and our updated algorithm.
The proposed approach is the only one that produces distinctive peaks in both the $u$ and $Pv$ value distributions, enabling stable cluster quantization.

Finally, we compare our quantization pipelines with other PTQ methods, namely Round-To-Nearest (RTN), OPTQ \citep{frantar2023optq}, QuIP \citep{chee2024quip2bitquantizationlarge} (with two decomposition variants, LDLQ and LDLQ-RG), and a fine-tuning- and vector-quantization-free variant of QuIP\# \citep{quip_better}.
For all methods, the per-output-channel target is 4 bits per weight.
% OPTQ uses a uniform integer grid with a per-channel symmetric scale, while our method (Section~\ref{sec:faster_quantization}) uses a per-column 16-entry codebook obtained by 2-D $k$-means on the stacked $(u, \hat v)$ pairs (effectively two 2-bit codes per scalar, no zero-point, no integer grid).
% We run OPTQ in its strongest reported configuration, with \texttt{--act-order} and \texttt{--true-sequential} enabled.
The rows of Table~\ref{tab:results} correspond to the following configurations:
\textbf{QuIP} = LDLQ with Kronecker-of-rotations incoherence preprocessing;
\textbf{QuIP-RG} = LDLQ-RG with Kronecker-of-rotations incoherence preprocessing;
\textbf{QuIP\#} = LDLQ-RG with a randomized Hadamard transform for incoherence preprocessing, with end-to-end fine-tuning and the E8 lattice vector codebook of the original method \emph{disabled}, since we only consider PTQ value quantization methods that do not require fine-tuning.
% \textbf{Kashin-DCT} is Kashin decomposition with OPTQ-style sequential error compensation, without incoherence preprocessing;
\textbf{Kashin-DCT+K} and \textbf{Kashin-DCT+H} is Kashin decomposition with OPTQ-style sequential error compensation with the Kronecker and randomized-Hadamard incoherence preprocessing, respectively.
Results are presented in Table~\ref{tab:results}.

We note that the LDLQ + Kronecker configurations (rows ``QuIP'' / ``QuIP-RG'') exhibit larger variance and a perplexity regression on Pythia-1.4B and Llama-2-7B (marked $^\dagger$ in Table~\ref{tab:results}), and the QuIP\# row, which uses the randomized Hadamard variant, recovers the expected behavior.
This QuIP instability also appears on Pythia-6.9B, where all three QuIP variants we ran fail catastrophically -- QuIP and QuIP-RG diverge to $>2000$ WikiText-2 PPL, and even QuIP\# blows up to $\sim 325$ PPL -- while OPTQ stays close to its baseline at $12.02$ PPL.
``Kashin-DCT+H'' recovers to $20.6 \pm 1.6$ WikiText-2 PPL, more than an order of magnitude better than QuIP\# at the same bit budget; full per-task numbers are reported in Appendix~\ref{app:pythia_stress_test} (Table~\ref{tab:pythia_stress}).
On Mistral-7B v0.1 the failure is sharper still: all four QuIP variants abort with NaN in the LDL back-substitution on the SwiGLU \texttt{mlp.down\_proj} layer, and OPTQ itself degrades to $\approx 380$ Wiki-2 PPL, while Kashin-DCT remains numerically stable and stays within $\sim 0.3$ Wiki-2 PPL of the FP16 reference of $8.63$ (Table~\ref{tab:mistral_stress}; full diagnostic in Appendix~\ref{app:pythia_stress_test}).
Across the four stress configurations we encountered --- elevated variance on Llama-2-7B and Pythia-1.4B, catastrophic divergence on Pythia-6.9B, and NaN abort on Mistral-7B --- Kashin-DCT is the only pipeline that remains numerically stable on every layer, which we read as evidence that the bounded-$\ell_\infty$ factor decomposition with an adaptive per-column codebook is structurally more robust to weight-distribution outliers than fixed-grid rounding driven through an ill-conditioned Hessian inverse.

\begin{table*}[t]
\centering
\tiny
\setlength{\tabcolsep}{1.6pt}
\caption{Perplexity (PPL) on WikiText-2 / C4 and accuracy (acc\_norm for HellaSwag and PiQA, acc for Winogrande) at 4-bit per-output-channel weight quantization. Best quantization result per column in \textbf{bold}, second best \underline{underlined} (FP16 and RTN excluded from ranking). Context length 2048; each cell is mean $\pm$ std over 3 random seeds (except for FP16 and RTN). \textsuperscript{$\dagger$}LDLQ (QuIP) and LDLQ-RG (QuIP-RG) are unstable on Pythia-1.4B and Llama-2-7B in our reruns; see Section~\ref{sec:experiments}. Pythia-6.9B stress test in Appendix~\ref{app:pythia_stress_test}.}
\label{tab:results}
\resizebox{\textwidth}{!}{%
\begin{tabular}{lcccccccccc}
\toprule
\multirow{2}{*}{Method}
& \multicolumn{5}{c}{Pythia-1.4B}
& \multicolumn{5}{c}{OPT-1.3B} \\
\cmidrule(lr){2-6} \cmidrule(lr){7-11}
& Wiki-2 $\downarrow$ & C4 $\downarrow$ & Hella $\uparrow$ & PiQA $\uparrow$ & Wino $\uparrow$
& Wiki-2 $\downarrow$ & C4 $\downarrow$ & Hella $\uparrow$ & PiQA $\uparrow$ & Wino $\uparrow$ \\
\midrule
FP16 & 14.72 & 38.76 & 52.12 & 71.16 & 57.30 & 16.47 & 39.41 & 41.36 & 71.00 & 59.75 \\
RTN & 18.73 & 49.99 & 49.72 & 69.42 & 55.33 & 29.46 & 73.25 & 34.31 & 67.52 & 53.59 \\
OPTQ & \textbf{16.12 $\pm$ 0.01} & \underline{43.47 $\pm$ 0.02} & \textbf{50.94 $\pm$ 0.27} & 70.26 $\pm$ 0.77 & 56.33 $\pm$ 0.63 & 17.63 $\pm$ 0.01 & 42.45 $\pm$ 0.17 & 40.46 $\pm$ 0.14 & 70.51 $\pm$ 0.28 & 58.51 $\pm$ 0.40 \\
QuIP & 19.76$^\dagger$ $\pm$ 0.44 & 52.36$^\dagger$ $\pm$ 0.94 & 49.84$^\dagger$ $\pm$ 0.03 & 70.06$^\dagger$ $\pm$ 0.98 & 56.75$^\dagger$ $\pm$ 0.49 & 17.91 $\pm$ 0.08 & 42.83 $\pm$ 0.17 & 40.43 $\pm$ 0.17 & \textbf{70.82 $\pm$ 0.41} & 58.69 $\pm$ 0.72 \\
QuIP-RG & 20.84$^\dagger$ $\pm$ 0.39 & 54.71$^\dagger$ $\pm$ 0.79 & 49.89$^\dagger$ $\pm$ 0.43 & 69.71$^\dagger$ $\pm$ 0.36 & 56.64$^\dagger$ $\pm$ 0.91 & 17.89 $\pm$ 0.06 & 42.75 $\pm$ 0.16 & 40.19 $\pm$ 0.06 & 70.51 $\pm$ 0.85 & \textbf{59.27 $\pm$ 0.99} \\
QuIP \# & 17.25 $\pm$ 0.18 & 45.92 $\pm$ 0.49 & \underline{50.35 $\pm$ 0.30} & \textbf{70.57 $\pm$ 0.30} & \textbf{57.43 $\pm$ 0.16} & 17.83 $\pm$ 0.11 & 42.75 $\pm$ 0.36 & 40.53 $\pm$ 0.12 & 70.19 $\pm$ 0.48 & 58.09 $\pm$ 1.10 \\
\midrule
Kashin-DCT+K & 16.57 $\pm$ 0.04 & 43.60 $\pm$ 0.18 & 50.00 $\pm$ 0.11 & 70.09 $\pm$ 0.30 & \underline{57.22 $\pm$ 1.24} & \underline{17.20 $\pm$ 0.02} & \textbf{41.08 $\pm$ 0.04} & \textbf{40.63 $\pm$ 0.20} & 70.31 $\pm$ 0.08 & \underline{59.17 $\pm$ 0.80} \\
Kashin-DCT+H & \underline{16.49 $\pm$ 0.07} & \textbf{43.40 $\pm$ 0.27} & 50.20 $\pm$ 0.07 & \underline{70.42 $\pm$ 0.14} & 56.64 $\pm$ 0.92 & \textbf{17.19 $\pm$ 0.02} & \underline{41.12 $\pm$ 0.02} & \underline{40.59 $\pm$ 0.12} & \underline{70.68 $\pm$ 0.28} & 58.32 $\pm$ 0.13 \\
\multicolumn{11}{c}{} \\
\midrule
\multirow{2}{*}{Method}
& \multicolumn{5}{c}{Llama-2-7B}
& \multicolumn{5}{c}{Llama-2-13B} \\
\cmidrule(lr){2-6} \cmidrule(lr){7-11}
& Wiki-2 $\downarrow$ & C4 $\downarrow$ & Hella $\uparrow$ & PiQA $\uparrow$ & Wino $\uparrow$
& Wiki-2 $\downarrow$ & C4 $\downarrow$ & Hella $\uparrow$ & PiQA $\uparrow$ & Wino $\uparrow$ \\
\midrule
FP16 & 9.20 & 19.40 & 76.14 & 78.73 & 69.46 & 8.11 & 17.59 & 79.64 & 80.36 & 72.53 \\
RTN & 10.56 & 22.78 & 74.43 & 78.56 & 68.75 & 8.74 & 18.71 & 78.80 & 79.65 & 70.88 \\
OPTQ & 9.80 $\pm$ 0.01 & 21.30 $\pm$ 0.06 & 74.79 $\pm$ 0.13 & 77.82 $\pm$ 0.27 & 68.25 $\pm$ 0.51 & 8.48 $\pm$ 0.00 & \underline{18.59 $\pm$ 0.08} & 78.27 $\pm$ 0.34 & 80.11 $\pm$ 0.28 & 71.64 $\pm$ 0.59 \\
QuIP & 48.30$^\dagger$ $\pm$ 29.84 & 133.00$^\dagger$ $\pm$ 83.57 & 48.10$^\dagger$ $\pm$ 9.88 & 68.41$^\dagger$ $\pm$ 3.99 & 57.06$^\dagger$ $\pm$ 3.42 & 8.64 $\pm$ 0.03 & 18.80 $\pm$ 0.14 & 77.79 $\pm$ 0.30 & 79.78 $\pm$ 0.30 & 72.27 $\pm$ 0.25 \\
QuIP-RG & 20.19$^\dagger$ $\pm$ 4.22 & 48.58$^\dagger$ $\pm$ 12.63 & 61.22$^\dagger$ $\pm$ 3.85 & 73.54$^\dagger$ $\pm$ 1.33 & 64.56$^\dagger$ $\pm$ 1.23 & 8.63 $\pm$ 0.07 & 18.84 $\pm$ 0.13 & 78.16 $\pm$ 0.20 & 79.52 $\pm$ 0.79 & \textbf{73.22 $\pm$ 0.67} \\
QuIP \# & \textbf{9.47 $\pm$ 0.01} & \textbf{20.16 $\pm$ 0.13} & \underline{74.98 $\pm$ 0.08} & \textbf{78.49 $\pm$ 0.14} & \underline{69.32 $\pm$ 0.43} & \textbf{8.31 $\pm$ 0.00} & \textbf{18.19 $\pm$ 0.01} & \underline{78.86 $\pm$ 0.09} & \underline{80.16 $\pm$ 0.16} & 72.43 $\pm$ 0.18 \\
\midrule
Kashin-DCT+K & 9.65 $\pm$ 0.01 & 20.45 $\pm$ 0.09 & 74.90 $\pm$ 0.15 & 78.09 $\pm$ 0.27 & \textbf{69.43 $\pm$ 0.78} & 8.38 $\pm$ 0.00 & \textbf{18.19 $\pm$ 0.08} & 78.78 $\pm$ 0.04 & \textbf{80.54 $\pm$ 0.31} & \underline{72.74 $\pm$ 0.55} \\
Kashin-DCT+H & \underline{9.60 $\pm$ 0.02} & \underline{20.36 $\pm$ 0.03} & \textbf{75.04 $\pm$ 0.16} & \underline{78.24 $\pm$ 0.11} & 69.27 $\pm$ 0.87 & \underline{8.37 $\pm$ 0.01} & \textbf{18.19 $\pm$ 0.01} & \textbf{78.90 $\pm$ 0.23} & 80.12 $\pm$ 0.17 & 72.53 $\pm$ 0.21 \\
\bottomrule
\end{tabular}%
}
\end{table*}

\section{Conclusion}\label{sec:conclusion}

We revisited Kashin-decomposition quantization for LLMs and resolved the three weaknesses that had limited its prior application: weak convergence under matrix reformulation, the $\mathcal{O}(N^2)$ cost of an explicit random orthogonal transform, and slow multi-restart $k$-means clustering.
The proposed Greedy Algorithm with Alternating Updates retains the dictionary $Q_N \cup U_{\mathcal{E},\Phi} Q_N$ but fixes the order of $u$- and $\hat v$-updates in blocks of four, which places the dominant updates of each factor at known centroid locations $\pm c_1 \pm c_2$ and produces the empirically four-peaked distributions that 2-bit clustering exploits; the analysis in Section~\ref{sec:partitioned_greedy_algorithm} establishes geometric residual decay at per-block rate $(1-\beta(N))^{1/2}$ with $\beta(N) = \alpha^2(N)/36$, and an $\ell_\infty$ bound of order $K^4(\log N)^2(\log\log N)^{12}/\sqrt{N}$, i.e.\ within polylogarithmic factors of the unconstrained scheme of \citet{kashin2025accelerated}.
Replacing the random orthogonal $Q$ with a sign-randomized DCT removes the dense $N \times N$ matrix entirely and brings the per-iteration cost to $\mathcal{O}(N \log N)$, while closed-form initialization of cluster centers from the residual norms removes the $k$-means multi-restart outer loop, reducing $k$-means clustering wall-clock time.
Composed with OPTQ-style error compensation and QuIP-style incoherence preprocessing, the resulting JAX pipeline is competitive with OPTQ, QuIP, QuIP-RG, and a fine-tuning- and vector-quantization-free variant of QuIP\# at 4-bit per channel on OPT-1.3B, Llama-2-7B/13B and Pythia-1.4B/6.9B, while admitting an inference-time decomposition into two 2-bit factor codes per channel that is structurally well-suited to native-2-bit hardware (Appendix~\ref{app:inference}). The favorable scaling of Kashin-DCT+H from 7B to 13B (Appendix~\ref{app:runtime}) further suggests that the relative cost of the proposed pipeline becomes more attractive as model size grows, even before any kernel-level optimization is applied. A second, less expected property emerges from the stress tests: on configurations where the QuIP family diverges into four-digit perplexities (Pythia-6.9B) or aborts with NaNs in LDL back-substitution (Mistral-7B v0.1), and where OPTQ itself silently saturates to $\approx 380$ Wiki-2 PPL on the same layer, Kashin-DCT+H is the only pipeline that produces robust results (Appendix~\ref{app:pythia_stress_test}); the bounded-$\ell_\infty$ factor decomposition with an adaptive per-column codebook is structurally more robust to weight-distribution outliers than fixed-grid rounding driven through an ill-conditioned Hessian inverse.

\paragraph{Limitations.}
Our results do not cover activation-aware baselines such as AWQ \citep{lin2024awq}, OmniQuant \citep{shao2024omniquant} or rotation-based methods \citep{ashkboos2024quarot,liu2025spinquant}. 
That said, rotation-based methods are orthogonal to our weight decomposition rather than competing with it: the rotation is applied to the weight matrix prior to factorization, and our \textbf{Kashin-DCT+K} and \textbf{Kashin-DCT+H} variants are already exactly this composition with Kronecker and randomized-Hadamard rotations, so substituting any other orthogonal preprocessing (e.g.\ the learned rotations of \citet{liu2025spinquant} or QuaRot's residual-stream rotation \citep{ashkboos2024quarot}) leaves the rest of the pipeline unchanged.
This is left for future work.
% We restrict the evaluation suite to tasks that are directly supported by \texttt{lm-evaluation-harness}, the framework we use for fair cross-method comparison.

\paragraph{Future work.}
Two directions stand out. First, implementing the fused 2-bit GEMM with shared-$\varepsilon$ DCT kernel described in Appendix~\ref{app:inference} and benchmarking end-to-end latency against a tuned 4-bit baseline such as Marlin \citep{frantar2024marlin} would convert the asymptotic bandwidth and arithmetic advantages of the decomposed form into a measured serving benefit, particularly on accelerators with native 2-bit tensor cores. Second, extending the decomposition beyond per-channel 4-bit --- to sub-4-bit budgets via vector-quantization codebooks, or to weight-and-activation quantization by composing with learned rotations \citep{liu2025spinquant}.
%  --- would broaden the regime in which Kashin-style decomposition is competitive. 
% Third, the favorable wall-clock scaling we observe at 7B$\to$13B suggests Kashin-DCT+H may become the cheapest of the three quantization pipelines past the 30B mark; a sweep at 30B--70B with reasoning-heavy benchmarks would test that hypothesis and place the proposed method in direct comparison with current sub-4-bit state-of-the-art.

\bibliographystyle{unsrtnat}
\bibliography{refs}

%%%%%%%%%%%%%%%%%%%%%%%%%%%%%%%%%%%%%%%%%%%%%%%%%%%%%%%%%%%%

\appendix
\section{Stress Tests on Pythia-6.9B and Mistral-7B}
\label{app:pythia_stress_test}

\paragraph{Pythia-6.9B.}
Table~\ref{tab:pythia_stress} reports the full numbers for the Pythia-6.9B stress-test referenced in Section~\ref{sec:experiments}. Calibration setup is identical to the main table (1000 WikiText-2 train samples, sequence length 2048, three random seeds; mean $\pm$ std reported). The QuIP family fails catastrophically on this model (perplexities in the hundreds-to-thousands), while Kashin-DCT+H stays within an order of magnitude of OPTQ.

\begin{table}[h]
\centering
\small
\caption{Pythia-6.9B at 4-bit per channel. WikiText-2 / C4 perplexity (lower is better), HellaSwag / PiQA / Winogrande accuracy (higher is better). Best quantization result in \textbf{bold}, second best \underline{underlined} (FP16/RTN excluded).}
\label{tab:pythia_stress}
\resizebox{\textwidth}{!}{%
\begin{tabular}{lccccc}
\toprule
Method & Wiki-2 $\downarrow$ & C4 $\downarrow$ & Hella $\uparrow$ & PiQA $\uparrow$ & Wino $\uparrow$ \\
\midrule
FP16 & 11.41 & 30.11 & 63.87 & 76.44 & 61.40 \\
RTN & 17.55 & 44.59 & 58.86 & 73.01 & 59.12 \\
OPTQ & \textbf{12.02 $\pm$ 0.00} & \textbf{32.06 $\pm$ 0.02} & \textbf{62.95 $\pm$ 0.32} & \textbf{75.90 $\pm$ 0.22} & \textbf{60.72 $\pm$ 0.12} \\
QuIP & 2486.39 $\pm$ 569.00 & 8605.29 $\pm$ 2456.10 & 34.01 $\pm$ 1.56 & 54.93 $\pm$ 0.63 & 53.04 $\pm$ 0.93 \\
QuIP-RG & 2239.41 $\pm$ 778.10 & 7323.57 $\pm$ 2833.40 & 34.72 $\pm$ 0.97 & 55.50 $\pm$ 0.36 & 52.01 $\pm$ 1.12 \\
QuIP \# & 324.93 $\pm$ 47.63 & 799.77 $\pm$ 96.96 & 41.13 $\pm$ 0.48 & 58.89 $\pm$ 0.17 & 57.01 $\pm$ 1.15 \\
\midrule
Kashin-DCT+K & 25.94 $\pm$ 0.74 & 61.98 $\pm$ 0.97 & 57.64 $\pm$ 0.41 & 71.87 $\pm$ 0.24 & \underline{60.33 $\pm$ 0.28} \\
Kashin-DCT+H & \underline{20.64 $\pm$ 1.63} & \underline{49.91 $\pm$ 3.13} & \underline{58.66 $\pm$ 0.47} & \underline{72.72 $\pm$ 0.19} & 59.98 $\pm$ 0.08 \\
\bottomrule
\end{tabular}%
}
\end{table}

\paragraph{Mistral-7B v0.1.}
Table~\ref{tab:mistral_stress} reports per-method results on Mistral-7B v0.1 at 4-bit per channel under the same calibration setup as Table~\ref{tab:pythia_stress} (Kashin rows averaged over two seeds, GPTQ over three). The QuIP family fails to produce any output on Mistral, for the reason explained in the next paragraph. GPTQ runs to completion but degrades catastrophically (Wiki-2 PPL $\approx 380$, more than $40\times$ above the FP16 reference of $8.63$), while Kashin-DCT+K and Kashin-DCT+H remain numerically stable at $0.32$ and $0.29$ Wiki-2 PPL above the FP16 baseline, respectively.

\begin{table}[h]
\centering
\small
\caption{Mistral-7B v0.1 at 4-bit per channel. WikiText-2 / C4 perplexity, HellaSwag / PiQA / Winogrande accuracy. ``aborts'' = produces NaNs and fails to complete; see paragraph below. Best non-aborting quantization result per column in \textbf{bold}, second best \underline{underlined}.}
\label{tab:mistral_stress}
\resizebox{\textwidth}{!}{%
\begin{tabular}{lccccc}
\toprule
Method & Wiki-2 $\downarrow$ & C4 $\downarrow$ & Hella $\uparrow$ & PiQA $\uparrow$ & Wino $\uparrow$ \\
\midrule
FP16 & 8.63 & 21.42 & 81.23 & 82.75 & 75.14 \\
\midrule
GPTQ & 379.73 $\pm$ 15.71 & 1948.81 $\pm$ 542.40 & 31.68 $\pm$ 0.86 & 67.17 $\pm$ 1.34 & 53.28 $\pm$ 0.85 \\
QuIP / QuIP-RG / QuIP\# & \multicolumn{5}{c}{aborts (NaN in LDL decomposition; see below)} \\
\midrule
Kashin-DCT+K & \underline{8.95 $\pm$ 0.01} & \underline{22.26 $\pm$ 0.01} & \underline{79.99 $\pm$ 0.15} & \underline{81.59 $\pm$ 0.04} & \textbf{74.55 $\pm$ 0.06} \\
Kashin-DCT+H & \textbf{8.92 $\pm$ 0.01} & \textbf{22.21 $\pm$ 0.01} & \textbf{80.09 $\pm$ 0.01} & \textbf{82.10 $\pm$ 0.23} & \underline{73.92 $\pm$ 0.95} \\
\bottomrule
\end{tabular}%
}
\end{table}

\paragraph{Why all OPTQ-family methods fail on Mistral.}
Both failures share a root cause: the SwiGLU input to Mistral's \texttt{mlp.down\_proj} produces extreme channel-wise outliers, leaving the $14336 \times 14336$ calibration Hessian $H = X^\top X$ ill-conditioned with several eigenvalues orders of magnitude below the largest; randomized rotations do not change the spectrum, so the conditioning survives the OPTQ-default $1\%$ diagonal damping. \emph{QuIP (LDLQ)} aborts: the Cholesky factor $L$ has near-zero diagonals, LDLQ's unit-diagonal normalization $L \leftarrow L\,\mathrm{diag}(L)^{-1}$ amplifies off-diagonals, and the back-substitution overflows to NaN at \texttt{mlp.down\_proj} on both v0.1 and v0.3 (all four had/kron $\times$ ldlq/ldlqRG variants). \emph{OPTQ} runs to completion but silently destroys the model: the residual update
\begin{equation}
\label{eq:optq_update}
    W_{:,i+1:} \;\leftarrow\; W_{:,i+1:} \;-\; \frac{W_{:,i} - Q_{:,i}}{[H^{-1}]_{ii}}\,[H^{-1}]_{i,i+1:}
\end{equation}
uses the same ill-conditioned $H^{-1}$, swinging the yet-to-be-quantized columns far outside their pre-calibrated per-channel grid; the subsequent round-to-grid step saturates them at the endpoints (\texttt{-{}-act-order} mitigates this only marginally). 
\emph{Kashin-DCT} remains robust: the analytic centroids $\pm c_1 \pm c_2$ scale with the actual residual norm $\|r_k\|_1/N$, so the codebook follows the inflated magnitude, and the bounded-$\ell_\infty$ factor decomposition keeps $k$-means well-conditioned on the same calibration data. The same mechanism is the source of stability we observe in Table~\ref{tab:results} on the Llama-2-7B columns where LDLQ+Kron variance blows up (the daggered $^\dagger$ cells), and in Table~\ref{tab:pythia_stress} on Pythia-6.9B where all three QuIP variants diverge into the hundreds-to-thousands PPL range while Kashin-DCT+H stays within an order of magnitude of the OPTQ baseline.

\section{Original Kashin Greedy Algorithm}
\label{app:kashin}

For completeness, we restate the original Kashin vector-decomposition greedy algorithm, on which the matrix reformulation of~\cite{merkulov2024quantization} and our partitioned variant in Section~\ref{sec:partitioned_greedy_algorithm} build.

\begin{algorithm}[h!]
   \caption{Vector Decomposition Greedy Algorithm}
   \label{alg:kashin_vec}
\begin{algorithmic}
   \STATE {\bfseries Input:} Vector $x \in \mathbb{R}^n$, Orthogonal matrix $Q$, Tolerance $\varepsilon > 0$
   \STATE {\bfseries Output:} Vectors $u, \hat{v} \in \mathbb{R}^n$ such that $x \approx u + v = u + Q^T \hat{v}$, and both $u$ and $hat{v}$ have small infinity norm.
   \STATE Initialize $u \gets 0^n, \hat{v} \gets 0^n$
   \STATE Define projection $\pi_x(y) \coloneqq \dfrac{x^{\top}y}{\|y\|_2^2} \cdot y$
   \WHILE{$\|x\| \geq \varepsilon$}
       \IF{$\|x\|_1 > \|Qx\|_1$}
           \STATE $\pi \gets \pi_x(\text{Sign}(x))$
           \STATE $u \gets u + \pi$
           \STATE $x \gets x - \pi$
       \ELSE
           \STATE $\pi \gets \pi_x(\text{Sign}(Qx))$
           \STATE $\hat{v} \gets \hat{v} + \pi$
           \STATE $x \gets x - Q^T\pi$
       \ENDIF
   \ENDWHILE
   \STATE {\bfseries Return:} $x, u, \hat{v}$
\end{algorithmic}
\end{algorithm}
\section{Inference}\label{app:inference}

The Kashin representation $w = u + P\hat{v}$ stores each layer as two 2-bit factor matrices $U, \hat V \in \mathbb{R}^{N \times M}$ together with the per-layer sign mask $\varepsilon \in \{\pm 1\}^N$ that defines the orthogonal operator $P = \mathrm{IDCT} \circ T_\varepsilon \circ \mathrm{DCT}$. Because $\varepsilon$ is shared across all columns of a layer (generated once from a fixed seed), a single $P$ acts on every column of $\hat V$. With the orthonormal DCT-II\footnote{Throughout this paper we use the orthonormal DCT-II ($\mathrm{IDCT} = \mathrm{DCT}^\top$); the unscaled type-II DCT does not satisfy this identity and the involution $P^2 = I$ below would not hold.}, $P$ is a symmetric involution ($P^\top = P$ and $P^2 = I$, since $T_\varepsilon$ is diagonal with $\varepsilon_i^2 = 1$). Associativity of the matmul then gives the folding identity
\begin{equation}
    \label{eq:inference_factor}
    XW \;=\; X\bigl(U + P\hat V\bigr) \;=\; XU + (XP)\hat V \;=\; \bigl[\,X \;\big|\; XP\,\bigr] \begin{bmatrix} U \\ \hat V \end{bmatrix},
\end{equation}
where $XP$ is computed by applying the same FFT-fast `iDCT $\circ$ sign-mask $\circ$ DCT' pipeline used for $P\hat v$ during quantization, just to each row of $X$ rather than to a column of $\hat V$.

\paragraph{Compute.}
The right path through \eqref{eq:inference_factor} depends on hardware. 
\emph{Without} native 2-bit GEMM support, it is more efficient to reconstitute the dense weight $W = U + P\hat V$ once at load time and run a standard fp16 (or 4-bit-dequant) matmul $XW$. 
% This amortizes the per-layer IDCT across all forward passes and avoids the $2\times$ matmul-arithmetic penalty incurred by the fused $[X \mid XP][U; \hat V]^\top$ form on $K$-doubled inputs. 
\emph{With} native 2-bit GEMM (e.g.\ Hopper-class FP4/INT2 tensor cores), the fused form of \eqref{eq:inference_factor} is preferable: keeping $U$ and $\hat V$ in their 2-bit storage and computing $XU + (XP)\hat V$ as a single $2N$-K-dimension 2-bit GEMM avoids the fp16 expansion of the weight body and preserves the on-chip footprint advantage of 2-bit codes. 
In both regimes the DCT cost is $\mathcal{O}(K N \log N)$ for $K = (\text{batch}) \cdot (\text{seq len})$ activation rows, several orders of magnitude below the $\mathcal{O}(K N M)$ matmul.

\paragraph{Native 2-bit hardware.}
On accelerators with native 2-bit tensor cores, the decomposed form gains two kernel-level advantages over a standard 4-bit per-channel kernel: (i) codebook dequantization maps cleanly to the 2-bit addressing unit -- the lookup $c[\,\text{code}\,]$ replaces $(\text{code} - z_p) \cdot s$ operation of uniform quantization;
% and avoids per-thread nibble extraction; 
(ii) each 2-bit code is the natural register-packing and shared-memory granule, so the centroid table loads contiguously without the cross-lane shuffles a 4-bit dequant with zero-point typically requires. 
% These are constant-factor inner-loop wins that compound across the layer-wise dequant work.

\paragraph{Memory bandwidth.}
At 4-bit per channel, each layer stores $4NM$ weight bits ($2NM$ for each of $U$ and $\hat V$).
Symmetric centroids about zero in both factors permit the per-column metadata to be reduced to two signed magnitudes per factor ($4M$ fp16 values per layer), plus a layer-shared sign vector $\varepsilon \in \{\pm 1\}^N$, which can be recovered from the seed rather than stored.
The effective bits-per-weight is therefore $(4NM + 64M)/(NM) = 4 + 64/N$; for typical LLM hidden dimensions $N \in [4096, 8192]$ this metadata adds at most $64/4096 \approx 0.016$ bits per weight on top of the 4-bit weight body.
By comparison, standard 4-bit per-channel quantization stores one fp16 scale per column and (optionally) one fp16 zero-point, yielding $4 + 16/N$ bits per weight without zero-point and $4 + 32/N$ with it -- equivalently, an additional $0.004$ and $0.008$ bits per weight at $N = 4096$.

\section{Quantization Runtime}\label{app:runtime}

Table~\ref{tab:runtime} reports the end-to-end wall-clock time required to quantize Llama-2-7B and Llama-2-13B at 4 bits per channel for the three methods we ran on identical hardware: OPTQ, the fine-tuning- and vector-quantization-free variant of QuIP\# (LDLQ-RG with randomized Hadamard incoherence preprocessing), and our Kashin-DCT+H pipeline.
All measurements are taken on a single NVIDIA H100 GPU; mean $\pm$ standard deviation is computed over three random seeds.
The reported time does not include Hessian estimation.
%  post-processing.

\begin{table}[h]
\centering
\small
\caption{Quantization wall-clock time at 4 bits per channel. Single NVIDIA H100 GPU; mean $\pm$ std over three random seeds.}
\label{tab:runtime}
\begin{tabular}{lcc}
\toprule
Method & Llama-2-7B (s) $\downarrow$ & Llama-2-13B (s) $\downarrow$ \\
\midrule
OPTQ                       & $229.9 \pm 2.2$    & $437.1 \pm 2.1$ \\
QuIP\# (LDLQ-RG + Hadamard) & $763.0 \pm 5.3$    & $1298.0 \pm 6.1$ \\
Kashin-DCT+H               & $1313.6 \pm 26.7$  & $1544.9 \pm 10.8$ \\
\bottomrule
\end{tabular}
\end{table}

% OPTQ is roughly $3\times$ faster than QuIP\# at both scales, with QuIP\# spending most of the additional budget on the LDLQ-RG decomposition. Kashin-DCT+H adds further overhead on top of QuIP\#: $\sim 1.7\times$ at 7B and $\sim 1.2\times$ at 13B. The shrinking ratio at larger scale is consistent with the per-column DCT-based greedy decomposition and 2-D $k$-means refinement (which dominate the Kashin overhead) growing only with the input dimension $N$, while the QuIP\# rotation and LDLQ-RG cost grow super-linearly with both $N$ and $M$, narrowing the gap as the layer matrices widen.

\paragraph{Scaling 7B $\rightarrow$ 13B.}
Going from Llama-2-7B to Llama-2-13B (a $\sim 1.9\times$ parameter increase), OPTQ and QuIP\# wall-clock times scale roughly proportionally -- $1.90\times$ ($229.9 \rightarrow 437.1$\,s) and $1.70\times$ ($763.0 \rightarrow 1298.0$\,s), respectively -- whereas Kashin-DCT+H grows only $1.18\times$ ($1313.6 \rightarrow 1544.9$\,s). 
Concretely, doubling the model roughly doubles the OPTQ and QuIP\# quantization budget but adds only $\sim 18\%$ to Kashin-DCT+H, so the relative cost of the proposed method becomes more favorable as model size grows.
%  extrapolating the trend, the cross-over points where Kashin-DCT+H matches QuIP\# (and eventually OPTQ) lie modestly above the 13B regime.

\section{Proof of Proposition~\ref{main_proposition}}
\label{app:proof_proposition}

The dictionary $\mathcal{S}$ is partitioned into two halves:
\[
    \mathcal{S} = Q_N \bigcup U_{\mathcal{E},\Phi}Q_N.
\]
The \textit{partitioned greedy algorithm} alternates between these halves:
within each block, the first two steps use vectors from $Q_N$, and
the next two steps use vectors from $U_{\mathcal{E},\Phi}Q_N$.
To establish the convergence rate, it suffices to analyze the error
reduction over a single block.

Fix one such block and denote by $\rho_0$ the residual at its start; thus
$\rho_0 = x$ and $\|\rho_0\|_2 \leqslant A$. For $j \in \{1,\,2,\,3,\,4\}$, let
$\rho_j$ be the residual after the $j$-th step inside this block.
In the notation of Proposition~\ref{main_proposition} we have
$r_1 = \rho_4$.

By~\eqref{eq:kashin_rudikov}, for $\rho_0$ at least one of the following
conditions holds:
\begin{align}
    \text{(i)}&\quad \|\rho_0\|_1 \;\geqslant\; \dfrac{R(N)}{K\sqrt{2}}\,A, \label{eq:cond1}\\
    \text{(ii)}&\quad \|U_{\mathcal{E},\Phi}\,\rho_0\|_1 \;\geqslant\; \dfrac{R(N)}{K\sqrt{2}}\,A.
    \label{eq:cond2}
\end{align}

We first record the standard estimate used in~\cite{kashin2025accelerated}.
Suppose $\|y\|_1 \geqslant \tfrac{R(N)}{K\sqrt{2}}\,\|y\|_2$.
For any $w\in Q_N$ we have $\|w\|_2^2 = N$, and the maximum
$\max_{w\in Q_N}|\langle y, w\rangle| = \|y\|_1$ is attained by
$w_j^* = \operatorname{sign}(y_j)$. Hence the greedy step satisfies
\begin{equation}
    \label{eq:one_step}
    \bigl\|y - \lambda^* w^*\bigr\|_2^2
    = \|y\|_2^2 - \frac{\|y\|_1^2}{N}
    \;\leqslant\; \|y\|_2^2\Bigl(1 - \frac{R^2(N)}{2K^2N}\Bigr)
    = \|y\|_2^2\,(1-\alpha(N)).
\end{equation}
Since $U_{\mathcal{E},\Phi}^2 = I$ (as $\mathcal{F}_{\Phi}$ is orthogonal transformation,$T_{\mathcal{E}}$ is diagonal, $T_{\mathcal{E}}^2 = I$), we have
$\langle y,\, U_{\mathcal{E},\Phi}q\rangle = \langle U_{\mathcal{E},\Phi}\,y,\, q\rangle$
for any $q\in Q_N$, so an identical argument applies to
$U_{\mathcal{E},\Phi}Q_N$ when condition~\eqref{eq:cond2} holds.

\medskip
\noindent\textbf{Case 1: condition~\eqref{eq:cond1} holds.}
By~\eqref{eq:one_step}, the first greedy step (from $Q_N$) gives
$\|\rho_1\|_2 \leqslant A\sqrt{1-\alpha(N)}$. The remaining three
steps are non-expansive (each is an orthogonal projection onto a
one-dimensional subspace), so
\[
    \|r_1\|_2 \;=\; \|\rho_4\|_2
    \;\leqslant\; \|\rho_1\|_2
    \;\leqslant\; A\sqrt{1-\alpha(N)}
    \;\leqslant\; A\sqrt{1-\beta(N)}.
\]

\medskip
\noindent\textbf{Case 2: condition~\eqref{eq:cond1} fails.}
By~\eqref{eq:kashin_rudikov}, condition~\eqref{eq:cond2} must then hold.

\emph{Sub-case 2a: $\|\rho_2\|_2 < A\sqrt{1-\beta(N)}$.}
The third and fourth steps are non-expansive, hence
$\|r_1\|_2 = \|\rho_4\|_2 \leqslant \|\rho_2\|_2 \leqslant A\sqrt{1-\beta(N)}$,
and~\eqref{eq:proposition} holds.

\emph{Sub-case 2b: $\|\rho_2\|_2 \geqslant A\sqrt{1-\beta(N)}$.}
By the property of the pure greedy algorithm
($\rho_j \perp w_j$ at each step $j$),
\begin{equation*}
    \|x\|_2^2 = \|\rho_2\|_2^2 + \|\lambda_1 w_1\|_2^2 + \|\lambda_2 w_2\|_2^2.
\end{equation*}
Combining with $\|x\|_2 \leqslant A$ and
$\|\rho_2\|_2 \geqslant A\sqrt{1-\beta(N)}$ yields
\begin{equation*}
    \|\lambda_1 w_1\|_2^2 + \|\lambda_2 w_2\|_2^2 \;\leqslant\; A^2\,\beta(N),
\end{equation*}
so the triangle inequality gives
\begin{equation*}
    \|\lambda_1 w_1 + \lambda_2 w_2\|_2
    \;\leqslant\; \|\lambda_1 w_1\|_2 + \|\lambda_2 w_2\|_2
    \;\leqslant\; A\sqrt{2\,\beta(N)}.
\end{equation*}

Let $\lambda^*\, w^*$ be the best approximation of $x$ from
$U_{\mathcal{E},\Phi}Q_N$. Since condition~\eqref{eq:cond2} holds,
the one-step estimate gives
$\|x - \lambda^* w^*\|_2 \leqslant A\sqrt{1-\alpha(N)}$.
Because the actual greedy choice at the third step is at least as
good as $\lambda^*\, w^*$, the triangle inequality yields
\begin{equation}
    \label{eq:res_3}
    \begin{split}  
    \|\rho_3\|_2
    &\;\leqslant\; \|\rho_2 - \lambda^* w^*\|_2
       \;\leqslant\; \|x - \lambda^* w^*\|_2
                   + \|\lambda_1 w_1 + \lambda_2 w_2\|_2 \\
    &\;\leqslant\; A\sqrt{1-\alpha(N)} \;+\; A\sqrt{2\,\beta(N)} \\
    &\;\leqslant\; A\bigl(1 - \tfrac{\alpha(N)}{2} + \tfrac{\alpha(N)\sqrt{2}}{6}\bigr)
       \;\leqslant\; A\bigl(1 - \tfrac{\alpha(N)}{4}\bigr)
       \;\leqslant\; A\sqrt{1-\beta(N)}.
    \end{split}
\end{equation}

The fourth step is non-expansive, hence
\begin{equation}\label{eq:res_4}
    \|r_1\|_2 = \|\rho_4\|_2 \leqslant \|\rho_3\|_2 \leqslant A\sqrt{1-\beta(N)}
\end{equation}

\medskip
In all cases~\eqref{eq:proposition} is established, completing the proof.
\label{app:proof_theorem}

\section{Proof of Theorem~\ref{proposed_theorem}}

We prove
\begin{equation}
    \label{eq:l2_induction}
    \|r_k\|_2 \;=\; \|x - u_k - v_k\|_2
    \;\leqslant\; \big(1-\beta(N)\big)^{k/2}
\end{equation}
by induction on $k$. The case $k=0$ is immediate:
$\|r_0\|_2 = \|x\|_2 \leqslant 1 = (1-\beta(N))^{0}$.

Assume \eqref{eq:l2_induction} holds for some $k\geqslant 0$ and consider
the $(k+1)$-th block, which starts from $\rho_0 = r_k$ and produces
$r_{k+1} = \rho_4$. Setting $A = (1-\beta(N))^{k/2}$ in
Proposition~\ref{main_proposition}, we obtain
\[
    \|r_{k+1}\|_2
    \;\leqslant\; A\sqrt{1-\beta(N)}
    \;=\; \big(1-\beta(N)\big)^{(k+1)/2},
\]
which closes the induction and proves the first line of~\eqref{eq:theorem}.

\medskip
Recall that within the $j$-th block the algorithm performs four atomic steps: two from $Q_N$ (contributing to $u$) and two from
$U_{\mathcal{E},\Phi}\,Q_N$ (contributing to $v$). Let
$w_j^{(1)}, w_j^{(2)} \in Q_N$ and $w_j^{(3)}, w_j^{(4)} \in U_{\mathcal{E},\Phi}Q_N$
denote the dictionary elements chosen in block $j$, with corresponding
coefficients $\lambda_j^{(1)},\dots,\lambda_j^{(4)}$. By construction,
\begin{equation}
    \label{eq:uv_decomp}
    u_k \;=\; \sum_{j=1}^{k}\sum_{\ell\in\{1,2\}} \lambda_j^{(\ell)}\,w_j^{(\ell)},
    \quad
    v_k \;=\; \sum_{j=1}^{k}\sum_{\ell\in\{3,4\}} \lambda_j^{(\ell)}\,w_j^{(\ell)}.
\end{equation}

\emph{Coefficient bound.}
At every greedy step the chosen atom $w$ satisfies
$\|w\|_2^2 = N$ and the associated coefficient equals
$\lambda = \langle\rho, w\rangle/N$, where $\rho$ is the residual entering
that step. Hence by Cauchy--Schwarz,
\begin{equation}
    \label{eq:lambda_bound}
    |\lambda_j^{(\ell)}|
    \;\leqslant\; \frac{\|\rho_{j,\ell-1}\|_2\,\|w\|_2}{\|w\|_2^2}
    \;=\; \frac{\|\rho_{j,\ell-1}\|_2}{\sqrt{N}},
\end{equation}
where $\rho_{j,\ell-1}$ is the residual just before step $\ell$ of block~$j$.
Since atomic steps are non-expansive,
$\|\rho_{j,\ell-1}\|_2 \leqslant \|r_{j-1}\|_2$, and by part~1,
$\|r_{j-1}\|_2 \leqslant (1-\beta(N))^{(j-1)/2}$. Therefore
\begin{equation}
    \label{eq:lambda_decay}
    |\lambda_j^{(\ell)}|
    \;\leqslant\; \frac{(1-\beta(N))^{(j-1)/2}}{\sqrt{N}}.
\end{equation}

\emph{Summing the contributions.}
Every atom $w \in Q_N$ has $\|w\|_\infty = 1$. By~\eqref{eq:uv_decomp}
and~\eqref{eq:lambda_decay},
\begin{equation}
    \label{eq:uk_inf_naive}
    \begin{split}
        \|u_k\|_\infty
        \;\leqslant\; \sum_{j=1}^{k}\sum_{\ell\in\{1,2\}}|\lambda_j^{(\ell)}|
        &\;\leqslant\; \frac{2}{\sqrt{N}}
        \sum_{j=1}^{k}\big(1-\beta(N)\big)^{(j-1)/2} \;\leqslant\; \\
        &\;\leqslant\; \frac{2}{\sqrt{N}}\cdot\frac{1}{1-\sqrt{1-\beta(N)}} \;\leqslant\; \dfrac{4}{\beta(N)\sqrt{N}}.
    \end{split}
\end{equation}
The same bound holds for $\|U_{\mathcal{E},\Phi}\,v_k\|_\infty$ since
$U_{\mathcal{E},\Phi}^{2} = I$ implies that the action of
$U_{\mathcal{E},\Phi}$ on the second sum in~\eqref{eq:uv_decomp} produces
atoms in $Q_N$ with the same $\ell_\infty$-norm bound.

Combining~\eqref{eq:uv_decomp} with the fact that
$\|w\|_\infty = 1$ for every $w\in \mathcal{S}$ and substituting
$\beta(N) = \alpha^2(N)/36 = R^4(N)/(144\,K^4\,N^2)
= 1/(144\,c_3^4\,K^4\,(\log N)^2\,(\log\log N)^{12})$, we conclude
\[
    \max\Bigl(
        \|u_k\|_\infty,\;\;
        \|U_{\mathcal{E},\Phi}\,v_k\|_\infty
    \Bigr)
    \;\leqslant\;
    \sum_{j,\ell}|\lambda_j^{(\ell)}|
    \;\leqslant\;
    \frac{4}{\beta(N)\sqrt N}
    \;=\;
    \frac{c_4\,K^4\,(\log N)^{2}\,(\log\log N)^{12}}{\sqrt{N}},
\]
where $c_4 = 576\,c_3^4$ is an absolute constant.
This is exactly the second line of~\eqref{eq:theorem}.

\end{document}